\documentclass[letterpaper]{article}
\usepackage{iccc}
\pdfoutput=1

\usepackage{times}
\usepackage{helvet}
\usepackage{courier}
\usepackage{graphicx}
\usepackage{caption}
\usepackage{subcaption}
\usepackage{amsmath}
\usepackage{amsfonts}
\usepackage{booktabs}

\title{Shape Inference and Grammar Induction for Example-based Procedural Generation}
\author{Gillis Hermans, Thomas Winters and Luc De Raedt\\
Computer Science Department\\
KU Leuven, Belgium\\
gillis.hermans@student.kuleuven.be; \{thomas.winters, luc.deraedt\}@kuleuven.be}
\begin{document} 
\maketitle
\begin{abstract}
\begin{quote}
Designers increasingly rely on procedural generation for automatic generation of content in various industries.
These techniques require extensive knowledge of the desired content, and about how to actually implement such procedural methods.
Algorithms for learning interpretable
generative models from example content could alleviate both difficulties. 
We propose SIGI, a novel method for inferring shapes and inducing a shape grammar from grid-based 3D building examples.
This interpretable grammar is well-suited for co-creative design.
Applied to \emph{Minecraft} buildings, we show how the shape grammar can be used to automatically generate new buildings in a similar style.
\end{quote}
\end{abstract}

\section{Introduction}
Procedural modeling \cite{parish2001} and procedural content generation (PCG) \cite{yannakakis2018artificial} are used to co-creatively and automatically generate content for applications such as video games, films and simulations.
As the complexity and scope of these applications grows, these methods are increasingly relied upon to generate content.
Yet, the creation of procedural rules that can generate a particular type and style of content is a difficult and time-consuming process \cite{stava2010}.
Instead of creating rules by hand, it is possible to learn rules from examples.
Furthermore, one of the critical challenges of PCG is \emph{style inference}, or the ability to generate content \emph{in the same style} that has been learned or inferred from examples \cite{togelius_et_al:DFU:2013:4336}.
Solving these challenges could also be considered a step in the direction of computational creativity \cite{toivonen2015}, as learning new styles is a crucial part of their further exploration.

In this paper, we propose SIGI \emph{(Shape Inference and Grammar Induction)}, a novel method for inferring the style of one or more grid-based 3D buildings in the form of a shape grammar \cite{stiny1980}.
This grammar of geometric designs defines the style as a set of building style features, such as columns and windows, and their relations, and can be used to generate new buildings in a similar style.
We make the following contributions:
\begin{itemize}
    \item
    We propose a method for inferring shapes present in grid-based 3D buildings. Unlike most previous work, shape inference allows the segmentation of examples with limited user input and without predefined feature classes.
    
    \item
     We show how these shapes are used to induce a shape grammar, which allows co-creative design and automatic generation of similar buildings in an interpretable way. Furthermore, SIGI allows for the induction of a shape grammar from multiple example buildings, which generate buildings in a shared style. Applied to \emph{Minecraft} examples, we demonstrate our approach and results.
     \item
     As a notion for repetition or symmetry in the examples, we define matching shapes that lead to an enlarged generative space and the generation of novel buildings.
\end{itemize}

\section{Shape Inference}
In this section, we introduce shape grammars and discuss shape inference, a component of SIGI that seeks a set of shapes that correspond to parts of the style present in the example buildings.
We further use \emph{style feature} as an informal notion for any building component, such as a window, wall, balcony or awning, that is part of the building style.

\subsection{Definitions}

\begin{figure*}[t]
    \centering
  \includegraphics[width=.8\textwidth]{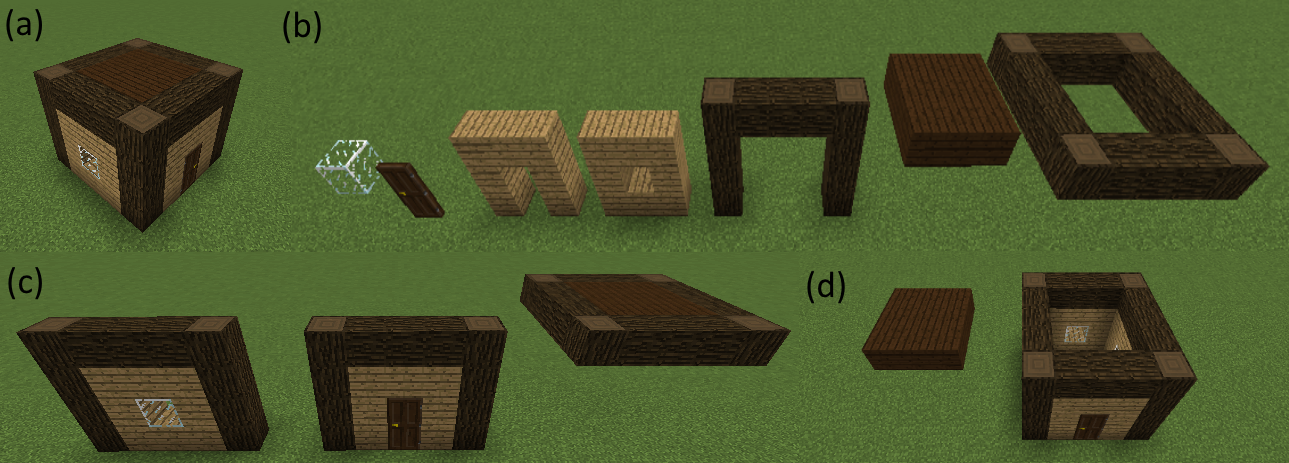}
  \caption{Example \textbf{(a)} and inferred shapes: 2D with $\alpha=0.75$ (see Equation \protect\ref{eq:basic}) \textbf{(b)}, rectangular with $\alpha=1.0$ \textbf{(c)} and 3D with $\alpha=0.25$ \textbf{(d)}. Inferred with merge operations and overlap. Shapes were rotated and matching shapes were combined for sake of clarity.}
  \label{fig:shapes}
\end{figure*}

Given grid-based 3D buildings $E$, SIGI infers shapes $S$ and induces a shape grammar $G$ for the style of $E$.
Input examples $E$ are composed of elementary components at a position in $E$.
We use \emph{Minecraft} buildings as examples, which exist out of a set of elementary blocks $B$.
A \emph{block} $b$ is a tuple $(t,p)$ with a type $t$ and a position $p = (x,y,z)$ in a grid-based 3D coordinate space in $\mathbb{Z}$.
A block represents a voxel at $p$ in a \emph{Minecraft} world textured according to its type $t$.

We adapt the \emph{shape grammar} formalism \cite{stiny1980} for grid-based shapes as a 4-tuple $\langle S,L,R,I \rangle$ where:
\begin{itemize}
    \item $S$ is a finite set of shapes
    \item $L$ a finite set of labels
    \item $R$ a finite set of \emph{shape rules} of the form $\alpha \rightarrow \alpha\beta$ where $\alpha$ and $\beta$ are labeled shapes $(S,L)$
    \item $I$ the initial labeled shape of the form $(S,L)$
\end{itemize}
A \emph{shape} $s$ in $S$ is a set of blocks $B_s$.
$B_s$ is a subset of blocks $B$ in an example $e$ in $E$: $B_s \subseteq B$.
Blocks $B_s$ are connected such that all blocks in $B_s$ are reachable from every other block in $B_s$ by following a path through adjacent blocks in $B_s$.
Thus, $s$ forms a coherent segment of $e$, without any disconnected blocks.
A \emph{labeled shape} contains auxiliary data, in the form of symbols.
A \emph{shape rule} consists of a transformation $\tau$ from one labeled shape to another and can take into account this labeled information.
Shape grammars function similarly to formal grammars or other production systems:
starting from $I$ and by applying rules, shapes are rewritten to produce new geometric shapes (as in Figure \ref{fig:shapegrammarillustration}).

We define three \emph{shape specifications} in order to further constrain the shapes: (a) \emph{3D shapes} without further limitations, (b) \emph{2D shapes} limited to a single plane by restricting all blocks to the same position on the $x$, $y$ or $z$ axis and (c) \emph{rectangular shapes} which further limit 2D shapes to a rectangular form.
Figure \ref{fig:shapes} shows shapes inferred for each specification.
As rectangular shapes suffice for two-dimensional facades \cite{teboul2013}, these may as well suffice for 3D buildings consisting of facades.
However, more complex parts of buildings, such as slanted roofs, might be difficult to describe with rectangular shapes.

\begin{figure}[t]
  \centering
  \includegraphics[width=0.5\textwidth]{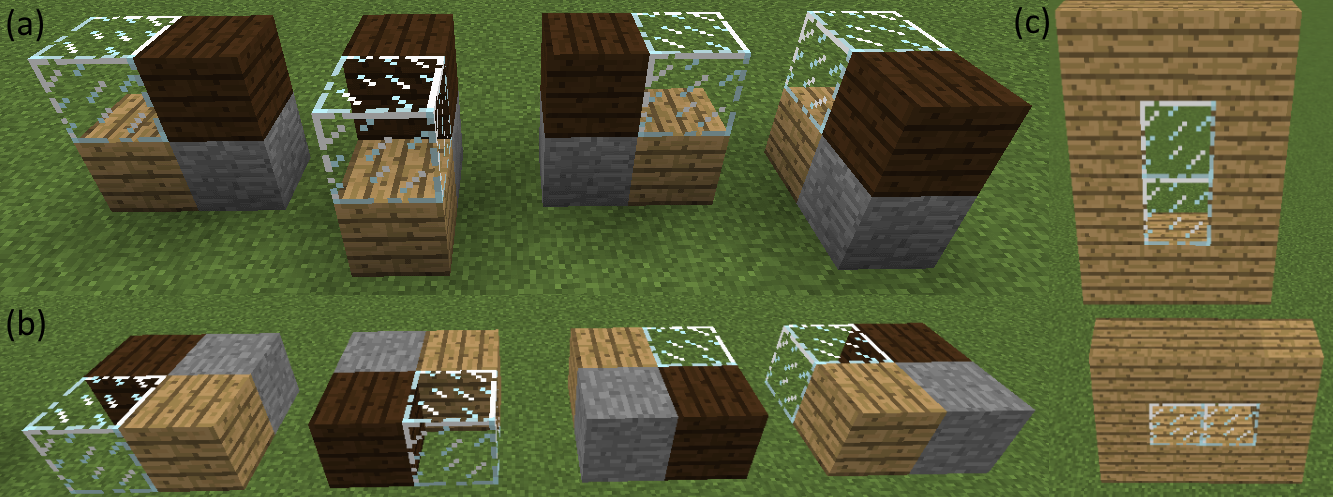}
  \caption{Matching vertical \textbf{(a)} and horizontal \textbf{(b)} shapes rotated along the z-axis. Two non-matching shapes with the same block configuration for which no $\tau_m$ exists \textbf{(c)}.}
  \label{fig:matchingrotation}
\end{figure}

We define \emph{matching shapes} as two shapes $s_i$ and $s_j$ that are identical, containing the same blocks in the same configuration, except for their position and orientation in the structure.
In other words, $s_i$ and $s_j$ \emph{match} if a transformation $\tau_m$ exists that is a one-to-one mapping of the blocks of $s_i$ onto $s_j$.
This transformation is of the form:
\begin{equation} \label{eq:transformation}
    \tau_m ( \begin{bmatrix}
        x_k\\
        y_k\\
        z_k
        \end{bmatrix} ) = R\begin{bmatrix}
        x_k\\
        y_k\\
        z_k
        \end{bmatrix} + \begin{bmatrix}
        \Delta x\\
        \Delta y\\
        \Delta z
        \end{bmatrix}
\end{equation}
where $\Delta x, \Delta y, \Delta z \in \mathbb{Z}$ and $R$ a rotation matrix along the $z$ axis (such that only rotations as in Figure \ref{fig:matchingrotation}\emph{(a,b)} are allowed):
\begin{equation}
    R = \begin{bmatrix}
        cos \theta & - sin \theta & 0 \\
        sin \theta & cos \theta & 0 \\
        0 & 0 & 1
        \end{bmatrix}
\end{equation}
with:
\begin{equation*}
    \theta \in \{0, \pm \frac{\pi}{2}, \pm \pi, \pm \frac{3 \pi}{2} \}
\end{equation*}
Matching shapes represent style features present in multiple locations on $E$, such as the rectangular window shapes in Figure \ref{fig:shapes} and \ref{fig:shapegrammarillustration} and the shared shapes in Figure \ref{fig:multiple}.

\subsection{Inferring Shape Sets}
In order to infer a suitable set of shapes $S$ for $E$ we apply a local search that minimizes a cost function on $S$.
Aside from ensuring shapes meet their requirements, we strive to find shapes that form a suitable description of the style.
Each shape $s$ ideally matches a style feature present in the examples.
Shapes are not limited to predefined feature classes, as in other work \cite{teboul2013,Martinovic_2013_CVPR}.
Instead, we infer a suitable set of shapes that are likely style features.

We make the assumption that most style features consist of a few components in a limited number of materials or block types.
A window, for example, usually exists out of glass and a frame in another material. While this assumption does not hold for any style feature, it establishes a foundation for the inference of shapes that are likely style features.
Thus we strive to find simple shapes (containing few block types), such that we avoid representing multiple style features in a single shape.
At the same time, we limit the total number of shapes, to avoid overly simple shapes devoid of any meaning.

\subsubsection{Cost function}
A suitable set of shapes consists of shapes that are neither too simple nor too complex.
We introduce a cost function that strives to find this balance.

Firstly, we limit the complexity of shapes by increasing their cost.
As a measure for this complexity, we use entropy $E_s$ \cite{shannon1951}, or the measure of information content, of a shape $s$:
\begin{equation} \label{eq:entropy}
    E_{s} = - \sum_{i=0}^{n} P(t_{i})log_{2}P(t_{i})
\end{equation}
where $n$ is the number of block types in $s$ and $P(t_{i})$ is the probability of block type $t_i$ in $s$.
The entropy cost favors compact and homogeneous shapes \cite{liu2011}, such as shapes consisting of just a single block.
To counterbalance this, we introduce a cost for the number of shapes $\#S$ in the set, adding by one to remove a bias for shape sets of size one.
By favoring a smaller number of shapes, the cost promotes larger shapes.
The resulting cost function to be minimized is:
\begin{equation} \label{eq:basic}
    (1+\#S)^{\alpha} \sum_{i=0}^{S} E_{s_{i}}
\end{equation}
where $s_{i}$ is a shape present in the shape set $S$ and $\alpha$ is a parameter weighing the importance of $\#S$.

\subsubsection{Local search}
The cost is minimized by repeatedly executing operations on $S$.
We use a hill-climbing algorithm that, at every step, evaluates and applies the first operation on the shape set that decreases the cost.
The algorithm converges to a local optimum, once no more operations exist that decrease the cost.
Figure \ref{fig:shapes} shows resulting shapes for different shape specifications and $\alpha$ values inferred from a simple example.

We define two operations on shapes.
A \emph{merge} combines $s_i$ and $s_j$ into one $s_n$, resulting in a new set of shapes $S'$:
\begin{equation*}
    S' = (S \backslash \{s_i,s_j\}) \cup \{s_n = \{s_i \cup s_j\} \}
\end{equation*}
A \emph{split} splits $s$ into two shapes $s_i$ and $s_j$:
\begin{equation*}
    S' = (S \backslash \{s\}) \cup \{s_i \subset s, s-s_i\}
\end{equation*}
Not all operations are legal, as resulting shapes must meet the defined requirements.
They must form coherent segments of $E$ and adhere to the chosen shape specification.

\begin{figure}[t]
  \centering
  \includegraphics[width=0.4\textwidth]{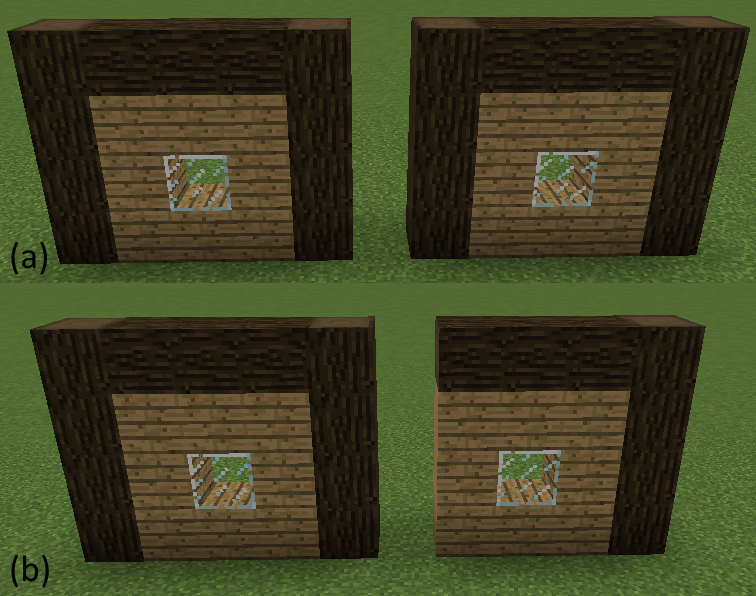}
  \caption{Two shapes inferred from the example in Figure \protect\ref{fig:shapes} with \textbf{(a)} and without overlap \textbf{(b)}. In this case more matching shapes are found with overlap.
  }
  \label{fig:overlap}
\end{figure}

During the execution of the local search, it is possible to use the merge, split or both operations.
These schemes require different initializations of $S$ at the start of the algorithm: minimal when merging, maximal when splitting and any in between for the combination.
When merging rectangular or 2D shapes, once two shapes have been combined they can no longer be separated.
One of the axes becomes fixed, locking the blocks out of potentially better shapes in other planes.
In order to alleviate this issue, we initialize each block as three shapes, with one in each plane, and only allow merges between shapes in the same plane.
After hill-climbing, we ensure all blocks in $E$ are present in $S$ and remove redundant shapes that are entirely covered by others.
A side effect of this optimization allows overlapping shapes, which can contain the same blocks, as in Figure \ref{fig:shapes}.
While the resulting shape set is no longer a pure segmentation of $E$, allowing overlapping shapes may increase the number of matching shapes, as illustrated in Figure \ref{fig:overlap}.
Some blocks may indeed belong to multiple style features present in the examples.

\begin{figure}[h]
  \centering
  \includegraphics[width=0.4\textwidth]{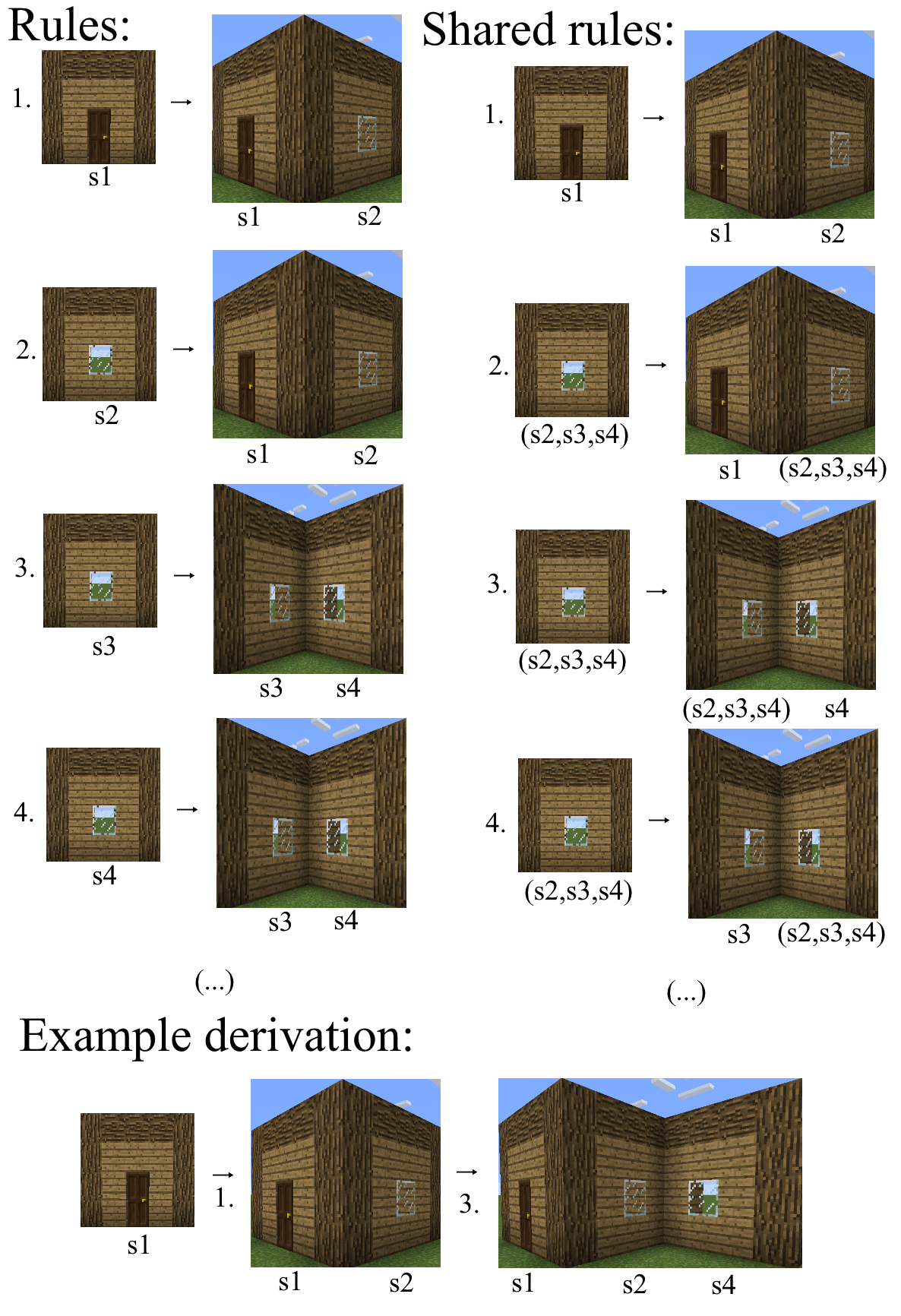}
  \caption{Partial shape grammar induced from the example and inferred rectangular shapes in Figure \protect\ref{fig:shapes}. This figure shows $4$ (of $16$) shape rules from Equation \protect\ref{eq:rule1}, the updated shared rules from Equation \protect\ref{eq:duperule1} and an example derivation of these rules. Matching shapes $(s_2,s_3,s_4)$ form shared rules that enlarge the grammar's generative space.}
  \label{fig:shapegrammarillustration}
\end{figure}

\section{Shape Grammar Induction}
We discuss how the inferred shape sets form a shape grammar, which can be used to produce new similar buildings.
Figure \ref{fig:shapegrammarillustration} shows a partial shape grammar induced by SIGI and an example derivation.

\begin{figure}[h]
  \centering
  \includegraphics[width=0.5\textwidth]{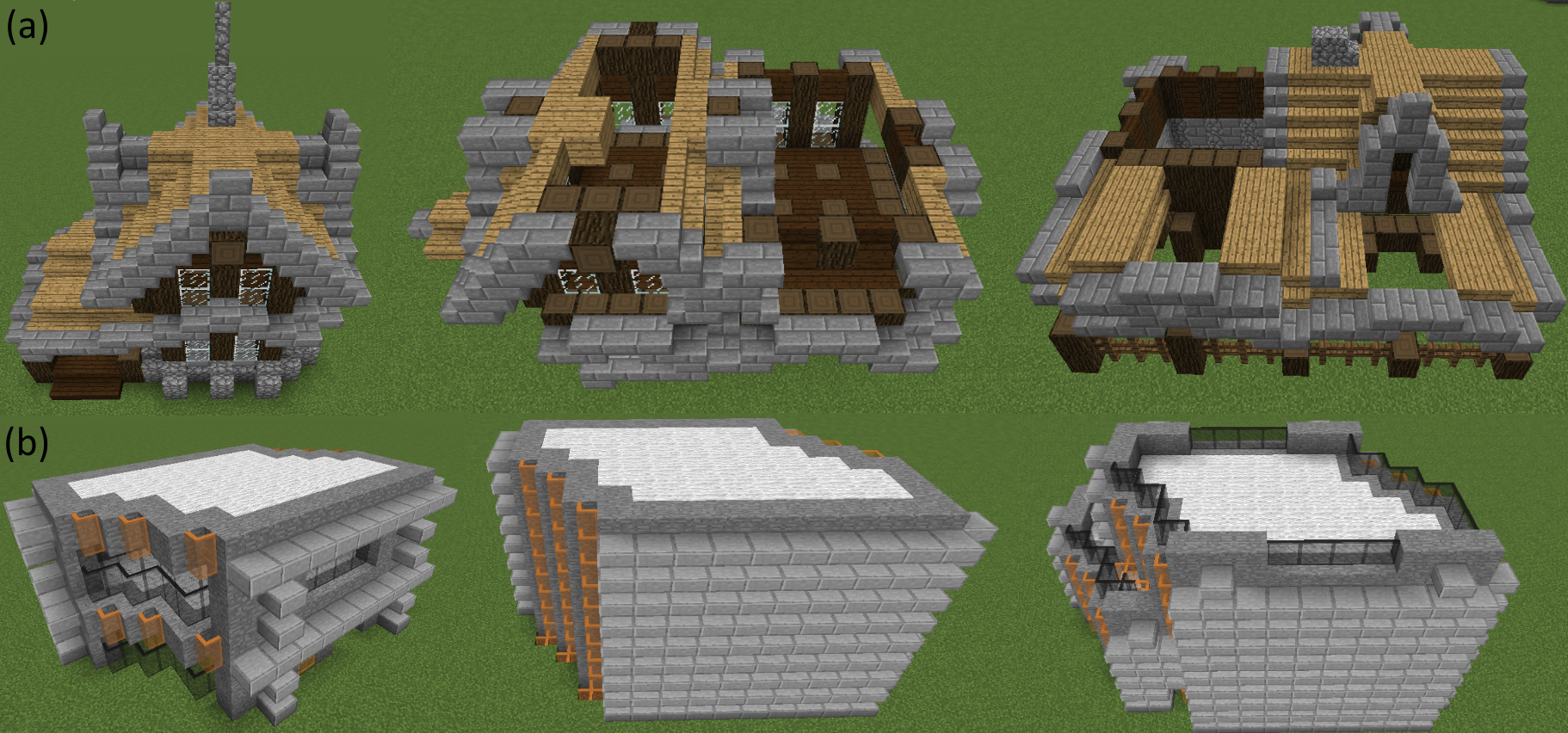}
  \caption{\textbf{(a)} Complex examples result in unstructured buildings. \textbf{(b)} Novel results from 2D shapes inferred as horizontal slices of the example.}
  \label{fig:extraexamples}
\end{figure}

\begin{figure*}[t]
  \includegraphics[width=\textwidth]{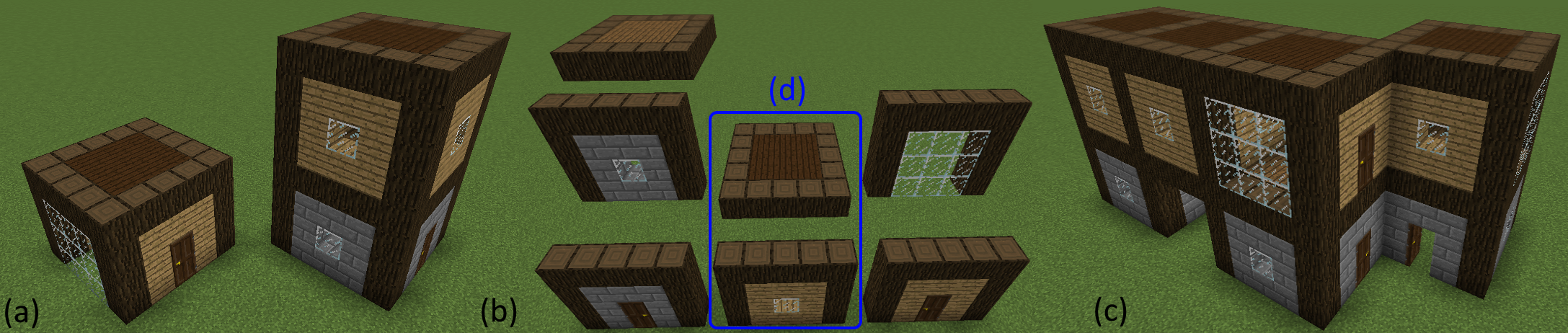}
  \caption{Two examples \textbf{(a)} with rectangular shapes \textbf{(b)} inferred with merges and $\alpha = 1.0$ and a new building \textbf{(c)} generated with the resulting shape grammar and the enclosure constraint. Shapes in the blue outline \textbf{(d)} are present in both examples and allow the combination of the two styles. Shapes were rotated and duplicate shapes were removed for sake of clarity.
  }
  \label{fig:multiple}
\end{figure*}

\subsection{Shape Rules}
Given the shapes $S$ for examples $E$, we induce a shape grammar $G$ with a set of shape rules $R$.
When a block $b_i \in s_i$ is directly adjacent to a block $b_j \in s_j$, these form two rules (as in Figure \ref{fig:shapegrammarillustration}):
\begin{equation}
    s_i \rightarrow s_i s_j \text{ and } s_j \rightarrow s_j s_i \label{eq:rule1}
\end{equation}
When the shape on the leftmost side of the rule is present in the production, the second shape can be added to the production.
Every shape is labeled with its original position and orientation in the example structure, such that during production of a shape it can be taken into account to calculate its new position.
A transformation $\tau_p$, of the same form as $\tau_m$ in Equation \ref{eq:transformation}, can be applied to the initial shape to move the shape to any other position.
When deriving a rule, if the leftmost shape has been transformed by $\tau_p$, the same transformation is applied to the rightmost shape.
Thus, the relative positions of both shapes in the example are retained in the production.
In this form $G$ generates just subsets of $E$.
In order to generalize the generative space of $G$ we make use of matching shapes in the shape set.

\paragraph{Shared rules}
While matching shapes represent the same style features in different positions in the examples, they form rules with different shapes.
We share rules between matching shapes, such that they can be applied to multiple shapes in $E$.
As an intuitive example, when a balcony is present next to a window in $E$, it can be produced next to any matching window shape in the production.
The production rules in Equation \ref{eq:rule1} extended to (as in Figure \ref{fig:shapegrammarillustration}):
\begin{equation}
    m_{s_i} \rightarrow m_{s_i}s_{j} \text{ and } m_{s_j} \rightarrow m_{s_j}s_{i} \label{eq:duperule1}
\end{equation}
where $m_{s}$ is the set of shapes that match $s$, including $s$.
These rules can be seen as shorthand for adding a duplicate rule for each shape that matches the leftmost shape in Equation \ref{eq:rule1}.
Starting from a shape $s$ in $m_{s_i}$ it is possible to expand a shared rule to add $s_j$ to the production.
The transformation $\tau_m$ in Equation \ref{eq:transformation}, that maps $s_i$ to its matching shape $s$, is applied to the new production $s_j$.
Thus, $s_j$ is transformed to form the same relative position with $s$ as was present in the example with $s_i$, as shown in the derivation of Figure \ref{fig:shapegrammarillustration}.
The sharing of rules between matching shapes allows rule expansions outside the space of $E$.
Thus, more matching shapes in $S$ generalize the generative space of $G$.
Additionally, if two buildings in $E$ have matching shapes, these two examples will be linked in $G$, because the matching shape rules provide a bridge between both production spaces.

\subsection{Production of Similar Structures}
The induced shape grammar $G$ allows the production of new artifacts in a similar style as the examples.
Starting from a production $P$, which contains the initial shape $I$ chosen from $S$, shape rules are applied that add shapes to $P$.
At every step the production selects a shape $s$ from $P$ and a rule $r$ that applies to $s$.
For all applicable rules the leftmost shape $s_l$ is either $s$ or a matching shape of $s$.
The rightmost shape $s_r$ is added to $P$ after applying $\tau_p$, if applied to $s_l$, and $\tau_m$, if $s$ is a matching shape of $s_l$.
These transformations align the relative positions of the shapes in the production.
A new shape is chosen from $P$, and the process can be repeated indefinitely.
When used as a co-creative tool, the designer controls the rule derivation and chooses a stopping point, both of which are hard to do sensibly automatically.

When using $G$ for automatic generation, rules are applied randomly until a predefined stopping condition, such as a maximum number of rule applications, is reached.
Shape grammars are not suited for automatic derivation, because unconstrained derivation frequently adds new shapes to the production \cite{wonka2003} and leads to unstructured buildings, as in Figure \ref{fig:extraexamples}$(a)$.
While a designer can guarantee structurally and creatively consistent artifacts by choosing which rules to expand, the quality of automatically generated artifacts is much more difficult to ensure.
Consequently, the derivations are usually done by hand or co-creatively with the assistance of a computer.
The automatic generation of satisfactory artifacts requires additional constraints, which are difficult to define in the shape grammar itself \cite{MERRICK2013115}. 

\paragraph{Enclosure constraint}
By removing redundant shapes or filling in empty spaces in an unstructured building, we form a coherent enclosed building with a clear distinction between the in- and outside.
We define the \emph{enclosure constraint} as follows.
Rectangular and 2D shapes are restricted to a single position on the $x$, $y$ or $z$ axis.
As such $s$ has two distinct sides $side_1$ and $side_2$ on either side of this fixed axis.
These consist of the positions of every block present in the shape shifted by $1$ or $-1$ along that axis.
The shape $s$ is enclosed when either $side_1$ or $side_2$ can not be reached through a path of empty space starting from the exterior of the structure.
We do not extend the enclosure constraint to 3D shapes, as there are no obvious sides to these shapes.
A simple pathfinding algorithm explores the production space and finds reachable sides in the production.
Once all sides have been explored, we remove any shape for which both sides were reachable, providing an enclosed production, as in Figure \ref{fig:multiple}.

\section{Evaluation and Results}
We implement\footnote{\label{github}https://github.com/gillishermans/sigi} SIGI as a filter for MCEdit-Unified\footnote{https://github.com/Podshot/MCEdit-Unified}, a world editor for \emph{Minecraft}.
Filters are written in Python code to extract and edit information from a \emph{Minecraft} world.
We evaluate the shape inference procedure through experiments on the effects of its parameters and perform a qualitative evaluation on the results of automatic generation.

\subsection{Shape Inference Evaluation}
SIGI provides a number of parameters for shape inference: three shape specifications and search operations, the $\alpha$ parameter and overlapping shapes.
The following experiments address the following question:
\begin{itemize}
    \item [Q] How do the inference parameters affect the results? 
\end{itemize}
As the \emph{ground truth} shape set of an example is undefined and ultimately comes down to the intentions of the designer, we emphasize objective summary measures instead of the correctness or value of the resulting shapes.
We perform shape inference on each example while alternating all parameter combinations and take aggregate measurements on the resulting shape sets.
These are: the number of shapes $\#S$, the percentage of matching shapes $\%M$, the number of blocks in a shape $Size$ and the number of block types over the shape size $C$ as a measure for complexity.
These allow us to estimate the average effects of the examined parameters in the composition of the shape sets.

\paragraph{Data}
The 9 examples\footnote{https://github.com/gillishermans/sigi\_results} used in these experiments
were chosen to encompass various complexities and structural features, such as slanted roofs and cylindrical buildings.
Examples $1-3$ were built for basic testing and $4-9$ were built by community members\footnote{https://www.planetminecraft.com/projects/} and edited to remove excess details.
Example building sizes range from $73$ to $498$ blocks and $5$ to $16$ different block types. 

\paragraph{Experiments}
As some blocks may belong to multiple style features in the examples, allowing overlap may increase the number of matching shapes.
As shown in Table \ref{table:overlap}, overlap increases matching shapes in general, but also significantly increases the number of shapes.
When considering the results of the examples side by side it seems overlap has a different effect on each example.
While effective for some examples, such as $E1$ in Figure \ref{fig:shapes},
it is detrimental for more complex examples with smaller shapes.
Thus, we do not recommend the use of overlap in general.

\begin{table}[ht]
\centering
\resizebox{\columnwidth}{!}{%
\begin{tabular}{@{\extracolsep{4pt}}llcccccccc}
\toprule   
{} & {} & \multicolumn{4}{c}{Average}  & \multicolumn{4}{c}{Median}\\
\cmidrule{3-6} 
\cmidrule{7-10} 
Examples & Overlap & \#S & \%M & Size & C & \#S & \%M & Size & C \\
\midrule
All & No & 27.23 & 27.8\%  & 96.33 & 0.15 & 6.0 & 21.9\% & 31.2 & 0.11 \\
{} & Yes & 51.04 & 31.2\% & 94.26 & 0.16 & 13.0 & 33.3\% & 22.68 & 0.13 \\
\midrule
E1 & No & 7.75 & 15.6\%  & 31.59 & 0.18 & 1.0 & 0.0\% & 25.0 & 0.08 \\
{} & Yes & 6.71 & 29.8\% & 33.80 & 0.13 & 5.0 & 40.0\% & 21.0 & 0.13 \\
\midrule
E5 & No & 33.05 & 28.1\%  & 137.14 & 0.14 & 26.0 & 8.8\% & 11.28 & 0.15 \\
{} & Yes & 72.53 & 26.9\% & 136.91 & 0.17 & 61.0 & 27.0\% & 9.18 & 0.16 \\
\bottomrule
\end{tabular}}
\caption{Results for \textbf{overlap} with the merge operation.}
\label{table:overlap}
\end{table}

The $\alpha$ parameter weighs the number of shapes in the cost function.
A higher value promotes a smaller set of shapes and thus larger shapes.
An $\alpha$ of $0.0$ produces minimal shapes because only entropy is taken into account.
Table \ref{table:alpha} shows that larger $\alpha$ lead to larger shapes and less matching shapes.
An $\alpha$ of around $5$ results in maximal shape sets, as the entropy is disregarded.
Thus any value in this range, specifically around $1.0$, will provide reasonably sized shapes for the shape set.

\begin{table}[ht]
\centering
\resizebox{\columnwidth}{!}{%
\begin{tabular}{@{\extracolsep{4pt}}lcccccccc}
\toprule
{} & \multicolumn{4}{c}{Average}  & \multicolumn{4}{c}{Median}\\
\cmidrule{2-5} 
\cmidrule{6-9} 
$\alpha$ & \#S & \%M & Size & C & \#S & \%M & Size & C \\
\midrule
0.0 & 64.60 & 52.0\% & 23.93 & 0.199  & 47 & 57.4\% &  6.60 & 0.168 \\
0.25 & 56.85 & 46.5\% &  43.82 & 0.165 & 36 & 51.2\% & 6.90 & 0.165 \\
0.5 & 55.29 & 41.7\% & 66.67 & 0.180 & 28 & 42.3\% & 9.04 & 0.152 \\
0.75 & 53.26 & 39.3\% & 68.84 & 0.177 & 24 & 40.0\% & 11.26 & 0.137 \\
1 & 42.85 & 33.2\% & 72.39 & 0.159 & 16 & 33.3\% & 19.27 & 0.127 \\
1.5 & 41.1 & 30.4\% &  78.00 & 0.157 & 12 & 30.0\%  & 22.90  & 0.124 \\
2 & 33.26 & 23.7\% & 100.28 & 0.144 & 8 & 22.2\% & 32.50 & 0.111 \\
5 & 31.04 & 23.6\% & 106.23 & 0.139 & 6 & 21.4\% & 35.00 & 0.109 \\
100 & 31.03 & 23.6\% & 106.27 & 0.139 & 6 & 21.4\% & 35.00 & 0.109 \\
\bottomrule
\end{tabular}}
\caption{Results for \textbf{\boldmath$\alpha$ parameter values}.}
\label{table:alpha}
\end{table}

The hill-climbing algorithm applies merge, split or the combination of operations.
Table \ref{table:operation} shows that the combination provides the most fine-tuned shape sets, as a merge and split can reverse each others effects.
Our implementation uses a minimal initialization for the combination as well as the merge.
Thus, it is more likely for these to converge soon, resulting in large shape sets with small shapes.
Conversely the split operation starts from maximal shapes and converges with larger shapes.
The merge and combination produce similar results because both start from the same initialization and follow the same initial path of merges.
Occasionally a split operation will occur in the combination, resulting in slightly larger shape sets with smaller shapes.
Consequently, the resulting shape set is highly reliant on the initialization of $S$, because the local search scheme converges quickly in the first local optima.

\begin{table}[ht]
\centering
\resizebox{\columnwidth}{!}{%
\begin{tabular}{@{\extracolsep{4pt}}lcccccccc}
\toprule 
{} & \multicolumn{4}{c}{Average}  & \multicolumn{4}{c}{Median}\\
\cmidrule{2-5} 
\cmidrule{6-9} 
Operation & \#S & \%M & Size & C & \#S & \%M & Size & C\\
\midrule
Merge & 47.1 & 34.2\% & 88.4 & 0.170 & 18.5 & 33.3\% & 18.3 & 0.148 \\
Split & 28.9 & 20.5\% & 106.54 & 0.120 & 5 & 16.7\% & 44.57 & 0.080 \\
Both & 41.54 & 34\% & 90.79 & 0.167 & 14 & 33.3\% & 16.43 & 0.147 \\
\bottomrule
\end{tabular}}
\caption{Results for \textbf{local search operations}.}
\label{table:operation}
\end{table}

SIGI allows three shape specifications: rectangular, 2D and 3D.
As shown in Table \ref{table:representation}, less constrained shapes allow larger shapes with less block types.
Even when paired with a low $\alpha$ (as in Figure \ref{fig:shapes}$(d)$), 3D shapes are often enormous, encompassing significant subsets of $E$ and lack matching shapes.
Thus, buildings generated from these shapes will lack variation.
We recommend a more fine-grained approach in the form of rectangular or 2D shapes.

\begin{table}[ht]
\centering
\resizebox{\columnwidth}{!}{%
\begin{tabular}{@{\extracolsep{4pt}}lcccccccc}
\toprule 
{} & \multicolumn{4}{c}{Average}  & \multicolumn{4}{c}{Median}\\
\cmidrule{2-5} 
\cmidrule{6-9} 
Shape & \#S  & \%M & Size & C & \#S  & \%M & Size & C \\
\midrule
Rectangular & 71.0 & 44.9\% & 12.10 & 0.261 & 43 & 42.3\% & 7.48 & 0.244 \\
2D & 38.5 & 36.5\% & 31.68 & 0.147 & 10 & 37.5\% & 29.13 & 0.124 \\
3D & 4.4 & 4.8\% & 256.10 & 0.037 & 1 & 0.0\% & 252.00 & 0.025 \\      
\bottomrule
\end{tabular}}
\caption{Results for \textbf{shape specifications}.}
\label{table:representation}
\end{table}

In conclusion, the combination of operations with rectangular or 2D shapes without overlap and an $\alpha$ between $0$ and $5$ generally result in the most suitable shape sets with fair number of matching shapes.

\subsection{Shape Grammar Evaluation}
We evaluate the shape grammar by means of a qualitative evaluation on the results of automatic generation, that aims to answer the following questions:
\begin{itemize}
    \item [Q1] Is SIGI able to induce a shape grammar capable of generating new buildings that are similar to the examples?
    \item [Q2] To what extent does SIGI infer the style of the example buildings?
\end{itemize}
Results were generated from the same examples used in the shape inference evaluation with two additional examples for a shared shape grammar.
We generated new buildings using $20$ and $50$ rule productions with and without enclosure.

While the results of an unconstrained automatic derivation of the shape grammar are unusable artifacts by themselves (Figure \ref{fig:extraexamples}$(a)$), enforcing the enclosure constraint can produce similar and suitable buildings (Figure \ref{fig:multiple} and \ref{fig:extraexamples}$(b)$).
Moreover, SIGI allows the induction of a shape grammar from multiple examples, and the generation of new buildings in a shared style (Figure \ref{fig:multiple}).
Results were found with straightforward shape inference parameters: $\alpha$ of $1.0$, rectangular or 2D shapes and merge or the combination of operations.
Although these could be tuned to further improve the results, at least for simple examples limited input is necessary.
Despite the success of Q1, our approach has a number of limitations.

\paragraph{Limitations}
SIGI struggles with buildings that include complex style features, such as slanted roofs, because they cannot be effectively represented with rectangular or 2D shapes.
Instead, these result in many small shapes and rules that complicate the shape grammar resulting in subpar productions (Figure \ref{fig:extraexamples}$(a)$).
Using 3D shapes leads to a few large shapes without any matching shapes (Figure \ref{fig:shapes}$(d)$).

The enclosure constraint is limited in a few ways.
Enclosure removes shapes even when unenclosed in the original examples.
A potential solution marks these during shape inference as unenclosed shapes which are ignored by the enclosure constraint.
At the same time it is possible that the removal of unenclosed shapes reveals new unenclosed shapes, as in Figure \ref{fig:extraexamples}$(b)$.
Thus, enclosure can be run multiple times until no shapes are removed.
Furthermore, there is no guarantee that any part of a generated building will be enclosed, resulting in empty generated artifacts.
Consequently this generation process is not suited for \emph{on the fly} generation, for example during gameplay.

This shape grammar considers only local relationships, in the form of adjacent shapes.
Our shape grammar is thus capable of extending shapes arbitrarily to form new structures in a similar style, which can result in novel buildings such as in Figure \ref{fig:multiple} and \ref{fig:extraexamples}$(b)$.
However, the style of buildings consists of a global structure \cite{mitra2014} in addition to the local structure, which SIGI does not take into account.
Thus as an answer to Q2, while SIGI is able to capture local style features and how they neighbor each other, the inferred style may be much more general than what we perceive as the style of a building due to neglecting the global structure.

\section{Discussion, Related and Future Work}
As an answer for the challenges of style inference \cite{togelius_et_al:DFU:2013:4336} and learning rules from examples \cite{stava2010}, we proposed SIGI, an approach towards shape inference and grammar induction from grid-based 3D buildings.
In this section we discuss SIGI, compare it to related example-based procedural methods and discuss future work.

\subsection{SIGI and Related Work}
\paragraph{Shape Inference}
SIGI employs a local search through candidate shape sets, minimizing a cost function, to infer shapes from examples.
This cost favors shapes that are likely style features, with the assumption that features contain a few block types.
Thus, shapes are not guaranteed to accurately represent style features, especially complex features for which this assumption does not hold.
In spite of this, the resulting grammar does not strictly require shapes that correspond tightly to style features.
Although these could improve the interpretability of the shape grammar, resulting in a more understandable derivation and modification process.
While resulting shapes are often satisfactory for simple examples, the search can get stuck in bad local optima.
Adding common local search refinements, such as backtracking and restarts \cite{aarts2003local} can alleviate this issue.
In SIGI each inferred set of shapes is limited to one shape specification.
However, style features are best represented by different types of shapes: rectangular shapes are ideal for walls but not for slanted roofs.
Thus, combining specifications (not exclusive to the ones defined in this paper) could better represent the examples.

Existing methods that induce shape grammars from 2D building facades rely on predefined feature classes \cite{teboul2013} and labeled input \cite{Martinovic_2013_CVPR}.
One approach that induces a grammar for 3D buildings \cite{aliaga2007} requires the user to subdivide the building into basic building blocks by hand.
However, model synthesis \cite{merrell2011} and inverse procedural modeling (IPM) \cite{bokeloh2010}, require limited user input in the form of a few parameters or constraints.
SIGI requires a similar amount of user input in the form of parameter choices, without any labeled data or predefined feature classes.

\paragraph{Shape Grammar Induction}
SIGI depends on matching shapes to enlarge the induced grammar's generative space and generate novel buildings.
Even when present in the examples, matching shapes must be found during shape inference.
A potential improvement redefines the cost function to encourage finding matching shapes.
Resulting shape grammars can be interpreted in a visual manner, as in Figure \ref{fig:shapegrammarillustration}.
Shape grammars are ideal for use in an interactive editor, similar to previously defined shape grammar interfaces \cite{muller2006,bokeloh2010}.
Such an editor allows co-creative design 
of new buildings by guiding the derivation and modifying the buildings and grammar.
The shape grammar is not inherently suited for automatic generation, because random application of rules adds new shapes to the production, without concern for the global structure \cite{wonka2003}.
Nonetheless, with the addition of enclosure, SIGI is capable of automatically generating suitable buildings for one or more simple examples.

Machine learning PCG methods, trained on example content, implicitly address the challenges of style inference and automatic rule learning \cite{pcgml}.
However, two common issues not present in SIGI are the lack of sufficient training data and the uninterpretable nature of many ML approaches.
Both model synthesis \cite{merrell2011} and IPM \cite{bokeloh2010}, the most similar approaches to SIGI, process 3D examples with limited user input and generate new structures in a similar style.
Just as SIGI, model synthesis allows the use of multiple examples towards a shared style.
Finding repetition or symmetry in the examples is an inherent problem in example-based PCG, which SIGI tackles with matching shapes and shared rules.
While model synthesis directly synthesises a new model from the examples by finding symmetric patterns, IPM induces a shape grammar by cutting the examples at symmetric parts.
Both methods rely on an \emph{adjacency constraint} (referred to as \emph{r-similarity} in IPM), which ensures that for every point $x$ in the generated model a point $x'$ exists in the example whose neighborhood matches the neighborhood of $x$.
This constraint, which could be applied to SIGI instead of enclosure, guarantees a local similarity within these neighborhoods defined by radius $\epsilon$.
Just as SIGI, these approaches produce similar structures by applying local relationships arbitrarily, without taking into account the global structure and style of the examples.

\subsection{Split Grammars}
It is possible to define constraints, such as the adjacency \cite{merrell2011,bokeloh2010} and enclosure constraint, that allow the shape grammar to generate suitable artifacts.
However, shape grammars are not inherently suited for automatic generation.
\emph{Split grammars} \cite{wonka2003} extend the shape grammar formalism in order to effectively use automatic rule derivation for generation.
They consist of a vocabulary of basic shapes (cuboids, cylinders, etc.) with split rules that split a basic shape into multiple other shapes and conversion rules that transform a basic shape into another.
In both cases the resulting shapes fit into the volume of the original shape.
The hierarchical nature of these rules constrains and controls automatic rule selection and ensures global structure.
In addition to the split grammar, that generates the basic structure, a control grammar distributes attributes, such as textures, throughout the structure.

\paragraph{Split Grammar Induction}
We propose an extension of SIGI for the induction of split grammars from examples.
Split grammars have previously been induced from two-dimensional examples \cite{Martinovic_2013_CVPR,teboul2013,muller2007}.
As far as we know, no example-based split
grammar approaches have been proposed for 3D structures.
One approach forms split rules by combining the shapes found during shape inference into larger shapes.
These are stripped of their block type data, thus specifying only block positions.
The shapes are combined again and again until a single shape remains, forming a top-down hierarchy of split rules, ideal for automatic generation.
Starting from the root shape, split rules are applied until the final shapes containing block type data are filled in.
As the hierarchy imposes a stopping point at which the new production is complete, this grammar is suited for \emph{on the fly} generation. 
This split grammar will, however, only generate copies of the original example.
By sharing rules with matching shapes, as discussed for shape grammars, we enlarge the generative space.
However at the same time, this split grammar will be limited in the variation provided along the higher-order global structure of buildings.
Thus, it may be necessary to induce the split grammar from multiple example buildings, in order to generate new buildings with differing global structure.
Instead of filling in block type data through rules, the addition of a control grammar that distributes attributes over the productions would improve the spatial distribution of the style in an orderly fashion, for example disallowing floating doors as present in Figure \ref{fig:multiple}.

While further details must be discussed and implemented in future work, SIGI provides a basis for a 3D example-based split grammar approach.

\section{Conclusion}
A current challenge for procedural generation is the induction of rules for the style of example content \cite{stava2010,togelius_et_al:DFU:2013:4336}.
We proposed SIGI, an example-based procedural approach towards the generation of grid-based 3D buildings.
SIGI induces a shape grammar from inferred shapes that
allows co-creative design of similar buildings in an interpretable way.
Furthermore, with the addition of the enclosure constraint, it is capable of automatically generating suitable buildings, with a few limitations.
A crucial aspect of this challenge is finding repetition and symmetry in the examples that lead to generalizations in the induced generative space, which we tackle with matching shapes and shared rules.
Existing methods, including SIGI, do not take into account the global structure of buildings, although it can be considered a crucial part of their style.
This paper serves as a stepping stone towards the currently unexplored problem of inducing split grammars from 3D examples, an approach that does take global structure into account and is inherently suited for automatic generation.

\section{Acknowledgments}
This research was partially funded by KU Leuven Research Fund (C14/18/062) and the Research Foundation - Flanders (G097720N). Thomas Winters is a fellow of the Research Foundation Flanders (FWO-Vlaanderen).

\bibliographystyle{iccc}
\bibliography{sigi}

\end{document}